\pdfoutput=1   % tell arXiv to use pdflatex (a PDF figure is included)
\documentclass[11pt]{article}

\usepackage[preprint]{acl}

\makeatletter
\@ifpackageloaded{lineno}{%
  \@ifpackagelater{lineno}{2026/02/11}{% v5.6 or newer: bug may be present
    \@ifpackagelater{lineno}{2026/02/17}{}{% v5.8+ assumed fixed: do nothing
      \ifdefined\RemoveFromHook \RemoveFromHook{build/column/before}[lineno]\fi
      \AtBeginDocument{%
        \ifdefined\@LN@orig@makecol
          \def\@LN@makecol{%
            \@LN@orig@makecol
            \setbox\@outputbox\vbox{\boxmaxdepth\@maxdepth
              \protected@write\@auxout{}{%
                \string\def\string\@LN@column{\if@firstcolumn1\else2\fi}}%
              \box\@outputbox}%
          }%
        \fi
      }%
    }%
  }{}% pre-v5.6: no bug, do nothing
}{}% no lineno: do nothing
\makeatother
\usepackage{times}
\usepackage{latexsym}
\usepackage[T1]{fontenc}
\usepackage[utf8]{inputenc}
\usepackage{microtype}

\hypersetup{
  pdftitle={GatePrune: Data-free Pruning of Self-attention Layers via Query-Key Coupling},
  pdfauthor={Dhananjay Saikumar and Blesson Varghese},
}

\usepackage{amsmath,amsfonts,bm}

\def\eqref#1{equation~\ref{#1}}
\def\1{\bm{1}}

\DeclareMathAlphabet{\mathsfit}{\encodingdefault}{\sfdefault}{m}{sl}
\SetMathAlphabet{\mathsfit}{bold}{\encodingdefault}{\sfdefault}{bx}{n}

\usepackage{amsmath,amssymb,amsfonts}
\usepackage{booktabs}
\usepackage{graphicx}
\usepackage{algorithm}
\usepackage{algpseudocode}
\usepackage{xcolor}      % for \arrayrulecolor
\usepackage{colortbl}    % provides \arrayrulecolor for coloring table rules
\usepackage{wrapfig}
\usepackage{commath}
\usepackage{tabularx}
\usepackage{pifont}      % for check/cross marks
\newcommand{\cmark}{\ding{51}}
\newcommand{\xmark}{\ding{55}}
\usepackage{arydshln}    % optional if you want dashed lines
\usepackage{tikz}
\usepackage{pgfplots}
\pgfplotsset{compat=1.17}
\usepgfplotslibrary{groupplots}
\usepackage{caption}
\usepackage{array}
\usepackage{enumitem}
\usepackage{multirow}
\newcommand{\revised}[1]{\textcolor{black}{#1}}

\title{GatePrune: Data-free Pruning of Self-attention Layers\\via Query-Key Coupling}

\author{Dhananjay Saikumar \\
  School of Computer Science \\
  University of St Andrews, UK\\
  \texttt{ds304@st-andrews.ac.uk} \\\And
  Blesson Varghese \\
  School of Computer Science \\
  University of St Andrews, UK\\
  \texttt{blesson@st-andrews.ac.uk}
}

\begin{document}
\maketitle

\begin{abstract}
Although the Self-Attention mechanism is a primary contributor to the inference latency of large language models (LLMs), many Self-Attention layers can be removed with little loss in accuracy. We attribute this to the \emph{Attention Suppression Hypothesis}: during pre-training, deeper Self-Attention layers learn to suppress their outputs, making them redundant. Leveraging this insight, we propose \textbf{GatePrune}, a one-shot, data-free method that scores and removes redundant Self-Attention layers from pre-trained weights alone---with \emph{zero calibration data, zero forward passes, and no fine-tuning}. Free of the calibration-set dependence of data-driven methods, GatePrune can prune a 13B-parameter model in \textbf{30\,s} only on a CPU when the model does not fit in GPU memory. It removes up to 40\% of the Self-Attention layers at a cost of at most 1.5 percentage points in average zero-shot accuracy, while speeding up inference by $1.30\times$ and accelerating pruning by $\mathbf{1000\times}$.
\end{abstract}

% ================================
% Main content
% ================================
% !TEX root = ../Main.tex
\section{Introduction}

Large language models (LLMs) have facilitated advances in translation, code generation, and reasoning~\citep{few_shot_learners,LLama-v1, GPT-4}. However, modern LLMs contain billions to hundreds of billions of parameters, requiring significant computational infrastructure and energy to drive them, thereby increasing deployment and environmental costs~\citep{carbon_footprint}. 
To reduce these costs, prior work has compressed models along two axes: \emph{reducing precision}, by quantizing weights to 8 or 4 bits~\citep{GPTQ, AWQ}, and \emph{reducing model width} by weight pruning to remove redundant parameters within layers~\citep{SparseGPT, wanda, owl}. However, these approaches require specialized hardware with low-precision or sparse-matrix kernels to achieve tangible speed-ups. Meanwhile, the third axis, \emph{reducing model depth}, i.e., the number of layers, remains a relatively under-explored area.

Modern LLMs process tokens using \(N\) Decoder-only Transformer blocks~\citep{radford2018improving, few_shot_learners}, each comprising two layers: (i) Self-attention for \emph{token mixing} across the sequence, and (ii) an MLP for independent \emph{channel mixing} per token. Although Self-attention layers typically contain only half the parameters of their MLP counterparts (e.g., 67M vs.\ 135M in \textsc{LLaMA-7B}~\citep{LLama-v2}), Self-attention is the primary contributor to inference latency due to its \(O(n^2)\) computational complexity for sequence length \(n\). For example, Self-attention takes twice as long as the MLP at \(n \approx 3000\), and three times longer by \(n \approx 7000\). Therefore, pruning even a few Self-attention layers can reduce inference costs and latency, irrespective of the available hardware.

Recent literature on pruning has leveraged \emph{structural} redundancy to reduce model depth~\citep{unreasonable_ineffectiveness} instead of weight pruning. \textsc{ShortGPT}~\citep{shortgpt} computes the cosine similarity between each Transformer block’s input and output to identify and prune blocks that exhibit high cosine similarity. However, pruning even a few blocks results in a substantial accuracy drop, suggesting that block-level pruning is an unsuitable granular unit for effective pruning.
To address this, the literature has explored pruning within blocks at the per-layer level (Self-attention vs.\ MLP). Since Self-attention layers are more redundant than MLP layers, nearly a third of the Self-attention layers can be removed with negligible loss in accuracy~\citep{not_all_attention}. 
These works rely on \textbf{thousands of calibration tokens} and \textbf{multiple forward passes} to compute the cosine similarity, and their accuracy is sensitive to the choice of calibration dataset~\citep{williams2024calibration}, such as C4~\citep{C4}, CodeAlpaca, MathInstruct~\citep{MathInstruct}, or LIMA~\citep{Lima}. This dependence results in a pruned model that is biased towards the calibration distribution, making it less reliable on unseen, zero-shot tasks.

% In contrast, an ideal \emph{data‐free} criterion - one that inspects only the model’s learned weights - enables \emph{on‐demand pruning} without any external data, hyperparameter tuning, or dataset leakage.  

\begin{figure}[t]
\captionsetup{font=footnotesize,labelfont=footnotesize}
  \centering
  \includegraphics[width=\columnwidth]{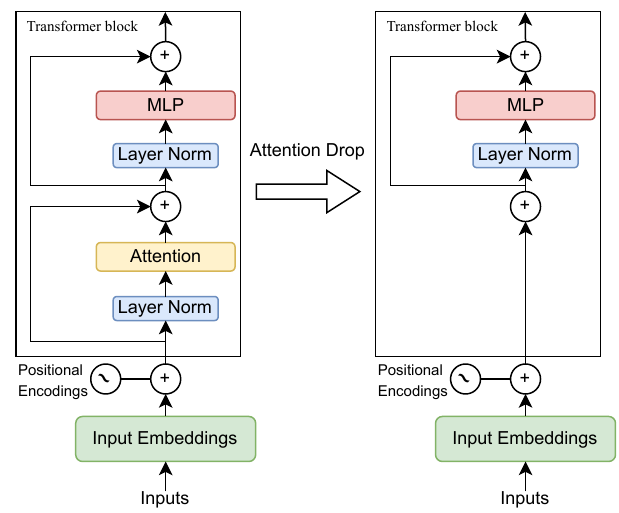}
  \caption{\textbf{Visualizing attention drop.}
           Removing the Self-Attention branch leaves a direct residual
           path feeding the MLP.}
  \label{fig:attn_drop}
\end{figure}

To eliminate the reliance on calibration data, this paper moves beyond empirical measurements and aims to understand the underlying mechanism at play. Therefore, we ask: \emph{why can many Self-attention layers be removed?} To answer this, we propose the \textbf{Attention Suppression Hypothesis}: during pre-training, deeper Self-attention layers learn to suppress their outputs, making them redundant. Consequently, these layers can be safely removed. Figure~\ref{fig:attn_drop} illustrates by pruning the residual stream bypasses Self-attention and feeds directly into the MLP.

We empirically validate the Attention Suppression Hypothesis by measuring the ratio of the Self-attention output to its input across layers, confirming that deeper layers suppress their output. Leveraging this insight, we introduce \emph{GatePrune}, a one-shot, data-free pruning method that identifies redundant Self-attention layers directly from pre-trained weights. Being data-free and forward-pass free, GatePrune operates in milliseconds on a GPU---over a \(\mathbf{1000\times}\) speed-up over data-driven methods---and can prune a 13B-parameter model in just \textbf{30 seconds} on a standard CPU using only system RAM, whereas calibration-based methods are prohibitively slow on CPUs.
 
GatePrune removes redundant Self-attention layers, offering comparable perplexity and zero-shot accuracy to the unpruned model, while speeding up inference by up to \textbf{1.3$\times$}.

In summary, our key contributions are as follows:
\begin{itemize}
  \item We formulate the \emph{Attention Suppression Hypothesis}, and validate that deeper Self-attention layers are redundant.
  \item We propose the first fully \emph{data-free} algorithm to prune Self-attention layers.
  \item We validate our approach on multiple LLMs, reducing inference latency while offering comparable perplexity and zero-shot accuracy to the unpruned model.
\end{itemize}

% !TEX root = ../Main.tex
\section{Related Work}

\subsection{Preliminaries and Notation}
We consider a decoder-only Transformer of depth \(L\), which processes an input token sequence ($X_{0}$) through \(L\) transformer blocks. At block index \(\ell\), the representation is \(
X_\ell \in \mathbb{R}^{B \times S \times D}\)
where \(B\) is batch size, \(S\) sequence length, and \(D\) hidden dimension. 

\paragraph{Self-attention layer:}  
This layer allows each token to ``attend'' to every other token in the sequence, performing \emph{token mixing} to aggregate contextual information across tokens. It is parameterized by trainable projection matrices \(W_{q,\ell}\), \(W_{k,\ell}\), and \(W_{v,\ell}\) for queries, keys, and values, as well as \(W_{o,\ell}\) for the output projection.
\begin{equation}
\begin{split}
Z_\ell &= \mathrm{Norm}(X_\ell), \\
\mathrm{AttnOut}_\ell &= \mathrm{MHA}(Z_\ell), \\
Y_\ell &= X_\ell + \mathrm{AttnOut}_\ell.
\end{split}
\label{eq:attn_block}
\end{equation}

Here, $\mathrm{Norm}$ is LayerNorm or RMSNorm (used by LLaMA), and multi-head attention (MHA) splits \(D\) into \(H\) heads of size \(d=D/H\). For each head \(h\):
% One definition per line: the original two-column alignat overflowed the
% column. Manual \tag{5}-\tag{7} also froze the equation counter, which risked
% these numbers being reused by later equations; auto-numbering avoids that.
\begin{align}
Q_\ell^{(h)} &= Z_\ell\,W_{q,\ell}^{(h)}, \\
K_\ell^{(h)} &= Z_\ell\,W_{k,\ell}^{(h)}, \\
V_\ell^{(h)} &= Z_\ell\,W_{v,\ell}^{(h)}, \\
L_\ell^{(h)} &= \tfrac{Q_\ell^{(h)}(K_\ell^{(h)})^\top}{\sqrt{d}}, \\
\Pi_\ell^{(h)} &= \mathrm{softmax}\bigl(L_\ell^{(h)}\bigr), \\
\mathrm{AttnOut}_\ell^{(h)} &= \Pi_\ell^{(h)}\,V_\ell^{(h)}\,W_{o,\ell}^{(h)}.
\end{align}
Concatenating across heads yields 
$
\mathrm{AttnOut}_\ell \in \mathbb{R}^{B\times S\times D}.
$

\paragraph{Feed-forward (MLP) layer:}  
Following Self-attention, each token’s representation is independently transformed by a two-layer MLP, performing \emph{channel mixing} without moving information across positions. This layer is parameterized by \(W_{1,\ell}\) and \(W_{2,\ell}\):
\begin{equation}
\begin{split}
U_\ell &= \mathrm{Norm}(Y_\ell), \\
\mathrm{MLP}_\ell &= W_{2,\ell}\,\sigma(W_{1,\ell}\,U_\ell), \\
X_{\ell+1} &= Y_\ell + \mathrm{MLP}_\ell.
\end{split}
\label{eq:resid_mlp}
\end{equation}

The MLP expands the token embedding to a higher‐dimensional hidden layer, applies an activation function, and projects back to dimension \(D\). This enables per‐token feature mixing and further refinement before passing to the next block.

\subsection{Cosine‐Similarity Importance}
To identify redundant blocks or layers, existing methods compute the cosine similarity between the input and the output of a given block~\citep{shortgpt, unreasonable_ineffectiveness}. Let $X_{\ell,b,t}$ and $X_{\ell+1,b,t}$ be the input and output of block $\ell$, respectively, for batch index $b$ and token $t$. The cosine similarity is defined as:
\begin{align}
\cos\bigl(X_{\ell,b,t},X_{\ell+1,b,t}\bigr)
&= \frac{X_{\ell,b,t}^\top X_{\ell+1,b,t}}
       {\|X_{\ell,b,t}\|\;\|X_{\ell+1,b,t}\|}, \label{eq:cosine}
\end{align}
which equals 1 when the block leaves its input unchanged. Averaging over all tokens ($T$ total) results in
\begin{align}
\bar c_\ell &= \frac{1}{T}\sum_{b,t}\cos\bigl(X_{\ell,b,t},X_{\ell+1,b,t}\bigr), \nonumber \\
\mathrm{Imp}_\ell^{\mathrm{block}} &= 1 - \bar c_\ell. \label{eq:block_imp}
\end{align}
Blocks with a low $\mathrm{Imp}_\ell^{\mathrm{block}}$ score have high similarity, indicating minimal representational change, and can therefore be removed.

\paragraph{Block Pruning:}
The ShortGPT framework~\citep{shortgpt} computes the importance score at a block level $\mathrm{Imp}_\ell^{\mathrm{block}}$ to rank and remove blocks. However, removing even 4 of 40 blocks in \textsc{LLaMA} 13B lowers average zero-shot accuracy by 1.7--3.9 percentage points for a speed-up of only $1.10\times$ (Table~\ref{tab:zero_shot_combined}). Meanwhile, removing 8 blocks yields a $1.25\times$ speed-up, but lowers accuracy by 9.2--11.2 points. Ultimately, block pruning quickly hits a wall of diminishing returns.

% ShortGPT treats each block as an atomic unit, but redundancy can be finer‐grained.

\paragraph{Layer Pruning:}
The limitations of block pruning have given rise to methods that operate at a finer granularity, pruning layers (Self-attention or MLP) within a block~\citep{not_all_attention}. To estimate the importance score at this level, the cosine similarity is computed between a layer's input and its combined output (the layer output and the residual stream). Specifically, these methods compare:
\begin{equation}
\begin{split}
Y_{\ell}^{\mathrm{attn}} &= \mathrm{AttnOut}_{\ell}+ X_{\ell}, \\
Y_{\ell}^{\mathrm{mlp}} &= \mathrm{MLP}_{\ell} + Y_{\ell}^{\mathrm{attn}}.
\end{split}
\label{eq:he_granularity}
\end{equation}
The importance score based on layer type is:
\begin{equation}
\begin{split}
\mathrm{Imp}_\ell^{\mathrm{attn}} &= 1 - \frac{1}{T}\sum_{b,t}\cos\bigl(X_{\ell,b,t},Y_{\ell,b,t}^{\mathrm{attn}}\bigr), \\
\mathrm{Imp}_\ell^{\mathrm{mlp}} &= 1 - \frac{1}{T}\sum_{b,t}\cos\bigl(Y_{\ell,b,t}^{\mathrm{attn}},Y_{\ell,b,t}^{\mathrm{mlp}}\bigr).
\end{split}
\label{eq:importance_scores}
\end{equation}

Empirical results across multiple LLMs~\citep{not_all_attention} demonstrate that Self-attention layers consistently yield lower importance scores than their MLP counterparts in deep layers ($\mathrm{Imp}_\ell^{\mathrm{attn}} \ll \mathrm{Imp}_\ell^{\mathrm{mlp}}$). A low importance score equates to high cosine similarity, meaning these Self-attention layers introduce minimal representational change and are highly redundant. Given that Self-attention is the primary computational bottleneck during inference, this difference in redundancy between the two types of layers directly motivates our focus on pruning Self-attention layers.

\subsection{Quantization, Weight Pruning, and Sparse Attention}
Prior methods have compressed models via quantization~\citep{GPTQ, AWQ, smoothquant} or weight pruning~\citep{SparseGPT, wanda, owl}, which often require low-precision or sparse-matrix kernels to reduce inference latency~\citep{Mosaic}. Sparse attention methods instead reduce the cost of each Self-attention layer by attending to only a subset of tokens, for example by keeping only the most important keys and values in the cache~\citep{h2o}. These methods reduce cost \emph{within} layers, whereas we remove entire Self-attention layers, which yields speed-ups without custom kernels and is complementary to them.

\section{The Attention Suppression Hypothesis}
\label{sec:active_suppression}

While prior work identifies redundant Self-attention layers via a cosine similarity-based importance score, the underlying mechanism for why many Self-attention layers are redundant remains unexplained. We propose the \textbf{Attention Suppression Hypothesis}: deeper Self-attention layers of an LLM learn to suppress their outputs.

The intuition behind the Attention Suppression Hypothesis is rooted in the Transformer architecture. In early transformer blocks, Self-attention performs \textbf{token mixing} to understand context, syntax and semantics. However, as representations become highly contextualized in deeper blocks, the network carries out \textbf{channel mixing} via the MLP to refine features for the next-token prediction. In the deeper blocks, token mixing becomes redundant, and thus, the Self-attention layers learn to suppress their outputs.

\subsection{Why Cosine Similarity Is Not Enough}

The cosine-based importance $\mathrm{Imp}_\ell^{\mathrm{attn}}$ (Equation~\ref{eq:importance_scores}) approaches zero for deep Self-attention layers~\citep{not_all_attention}. However, a cosine similarity near $1$ does not necessarily imply that the update $U=\mathrm{AttnOut}_\ell$ is small. Since cosine depends only on the direction of $X_\ell+U$, it is independent to any update parallel to the input. For example, $U$ $=cX_\ell$ with $c>0$ gives $\cos(X_\ell, X_\ell+U)=1$ for any magnitude. Suppression is instead evidenced by the attention-to-input norm ratio $r_\ell$ (Section~\ref{subsec:norm_ratio}, Figure~\ref{fig:attention-collapse}), which measures the full update. 

\subsection{Empirical Analysis of the Attention-to-Input Norm Ratio}
\label{subsec:norm_ratio}

To quantify attention suppression, we define the \textbf{attention-to-input norm ratio} $r_\ell = \frac{\sum \|\mathrm{AttnOut}_{\ell}\|}{\sum \|X_{\ell}\|}$. Since the input received by the MLP is the sum of the Self-attention layer output ($\mathrm{AttnOut}_{\ell}$) and the residual connection ($X_{\ell}$), $r_\ell$ measures the relative contribution of the Self-attention mechanism to the MLP's input.

As shown in Figure~\ref{fig:attention-collapse}, $r_\ell$ has three distinct phases across the model depth:
\begin{itemize}[leftmargin=*, noitemsep]
\item \textbf{High Influence:} In early layers, the norm of the Self-attention output is higher than that of the residual connection, indicating that token mixing dominates the representational change.
\item \textbf{Stable Plateau:} Across the middle layers, the ratio stabilizes, where the Self-attention mechanism contributes a consistent and proportional update to the residual connection.
\item \textbf{Attention Suppression:} In the later layers, the ratio exhibits a decay, with the Self-attention contribution vanishing to a negligible fraction of the residual connection.
\end{itemize}

This suggests that deeper Self-attention layers are largely redundant across multiple model families: removing them leaves the block output nearly unchanged (Appendix~\ref{app:block_output}).

Our observation is consistent with a well-studied line of work on \emph{attention rank collapse}, which shows that as Transformers grow deeper, the attention matrices of many layers degenerate towards rank-one structures~\citep{rank2,rank1,rank3}. The attention-to-input norm ratio $r_\ell$ offers a complementary, signal-level view of the same phenomenon: rank collapse characterizes the degeneracy of the attention \emph{map}, whereas $r_\ell$ quantifies the attention \emph{output} contributes to the residual stream, decaying sharply in the deep layers where collapse is reported. The combined evidence of rank collapse from the representation side and norm suppression from the signal side---further supports the Attention Suppression Hypothesis and motivates removing these layers.

\begin{figure*}[t]
  \captionsetup{font=footnotesize,labelfont=footnotesize}
  \centering
  \begin{tikzpicture}
    \begin{groupplot}[
        group style={
            group size=3 by 1,
            horizontal sep=0.8cm, 
        },
        % 3 panels + 2 separators must fit inside \textwidth
        width=0.37\textwidth,
        height=0.21\textwidth,
        xlabel={Layer Index},
        ymin=0, ymax=1.5,
        grid=major,
        grid style={dashed,gray!30},
        tick label style={font=\scriptsize},
        label style={font=\scriptsize},
        title style={font=\footnotesize\bfseries}
    ]
    
    % Plot 1: Vicuna-7B
    \nextgroupplot[
        title={Vicuna-7B},
        ylabel={$\|\mathrm{AttnOut}\| / \|X\|$},
        xmin=1, xmax=32,
        xtick={1,10,20,32},
        ytick={0,0.5,1.0,1.5}
    ]
    \addplot+[mark=*, thick, blue, mark size=1pt] coordinates {
        (1, 1.245) (2, 0.575) (3, 0.251) (4, 0.308) (5, 0.307) (6, 0.31) (7, 0.315) (8, 0.304) (9, 0.315) (10, 0.337) (11, 0.348) (12, 0.307) (13, 0.294) (14, 0.301) (15, 0.287) (16, 0.29) (17, 0.296) (18, 0.211) (19, 0.204) (20, 0.178) (21, 0.194) (22, 0.149) (23, 0.161) (24, 0.135) (25, 0.152) (26, 0.108) (27, 0.163) (28, 0.123) (29, 0.137) (30, 0.142) (31, 0.134) (32, 0.168)
    };

    % Plot 2: Vicuna-13B
    \nextgroupplot[
        title={Vicuna-13B},
        xmin=1, xmax=40,
        xtick={1,10,20,30,40},
        % yticklabels={,,} % Hides y-axis labels to save space
    ]
    \addplot+[mark=*, thick, blue, mark size=1pt] coordinates {
        (1, 1.427) (2, 0.52) (3, 0.441) (4, 0.463) (5, 0.315) (6, 0.316) (7, 0.313) (8, 0.331) (9, 0.334) (10, 0.328) (11, 0.341) (12, 0.333) (13, 0.332) (14, 0.322) (15, 0.32) (16, 0.316) (17, 0.284) (18, 0.259) (19, 0.225) (20, 0.217) (21, 0.216) (22, 0.218) (23, 0.152) (24, 0.139) (25, 0.13) (26, 0.124) (27, 0.122) (28, 0.101) (29, 0.111) (30, 0.096) (31, 0.094) (32, 0.095) (33, 0.089) (34, 0.086) (35, 0.095) (36, 0.094) (37, 0.126) (38, 0.158) (39, 0.154) (40, 0.19)
    };

    % Plot 3: LLaMA-13B v2
    \nextgroupplot[
        title={LLaMA-13B v2},
        xmin=1, xmax=40,
        xtick={1,10,20,30,40},
    ]
    \addplot+[mark=*, thick, blue, mark size=1pt] coordinates {
        (1, 1.38) (2, 0.48) (3, 0.44) (4, 0.46) (5, 0.32) (6, 0.31) (7, 0.32) (8, 0.33) (9, 0.33) (10, 0.34) (11, 0.33) (12, 0.32) (13, 0.32) (14, 0.31) (15, 0.31) (16, 0.30) (17, 0.28) (18, 0.25) (19, 0.22) (20, 0.21) (21, 0.20) (22, 0.22) (23, 0.14) (24, 0.13) (25, 0.12) (26, 0.12) (27, 0.11) (28, 0.10) (29, 0.09) (30, 0.08) (31, 0.08) (32, 0.07) (33, 0.07) (34, 0.07) (35, 0.07) (36, 0.08) (37, 0.10) (38, 0.13) (39, 0.15) (40, 0.16)
    };
    \end{groupplot}
  \end{tikzpicture}
  \caption{Attention-to-input norm ratio $r_\ell$ across layers for Vicuna-7B, Vicuna-13B, and LLaMA-13B v2.}
  \label{fig:attention-collapse}
\end{figure*}

% !TEX root = ../Main.tex
\section{Data-Free Pruning of Attention Layers Using GatePrune}
\label{sec:gate_norm}

Section~\ref{sec:active_suppression} demonstrates that during pre-training, deeper Self-attention layers learn to suppress their outputs, hence the majority of the signal is routed through the residual connection. We now introduce \textbf{GatePrune}: a data-free pruning method that identifies suppressed Self-attention layers by considering only pre-trained query and key weights. 

Unlike data-driven methods that require thousands of calibration tokens and multiple forward passes, GatePrune uses a weight-only importance score, \textbf{Gate-Norm}, to estimate the importance of each Self-attention layer, making it a data-free and forward-pass-free pruning algorithm.

\paragraph{Key Hypothesis:} 
\emph{If a layer's query--key coupling is weak, it cannot mix tokens selectively: its attention is near-uniform, and its token-selective contribution vanishes with the coupling.}

\subsection{Theoretical Rationale: From Weights to Bounded Updates}
We formalize the relationship between pre-trained weights and representational change in three steps (see Appendix~\ref{app:gate_norm_proofs} for the full derivation):

\paragraph{1. Gate-Norm Proxy:} 
For the $\ell$\textsuperscript{th} Self-attention layer, we define the \emph{gate matrix} $M_\ell = W_{q,\ell}W_{k,\ell}^\top$. The Frobenius norm, which is the \textbf{Gate-Norm}, upper-bounds the ``mixing strength'' that any two normalized token vectors can induce in the raw Self-attention logits:
\begin{equation}
m_\ell = \|M_\ell\|_F = \|W_{q,\ell}W_{k,\ell}^\top\|_F.
\label{eq:gate_norm}
\end{equation}
Notably, $M_\ell$ is the same query--key coupling whose norm governs the rate of attention rank collapse in prior theory~\citep{rank2,rank1}: we invert this \emph{diagnostic} relationship into a \emph{constructive}, data-free criterion for which Self-attention layers to prune.

\paragraph{2. From Gate-Norm to Uniform Attention:} 
The raw Self-attention logits are bounded by the Gate-Norm: $|L_{ij}| \le C_0 m_\ell$, where $C_0$ depends only on the normalized token norms. As $m_\ell \to 0$, the Self-attention mechanism approaches a \textbf{uniform distribution} over the sequence length $S$:
\begin{equation}
A_{ij} = \frac{e^{L_{ij}}}{\sum_k e^{L_{ik}}} = \frac{1}{S}\bigl(1 + O(m_\ell)\bigr).
\label{eq:softmax_uniform}
\end{equation}
% Deviations from exact uniform mixing ($\frac{1}{S}$) are bounded linearly by the Gate-Norm.
This near-uniform regime is consistent with the \emph{attention rank collapse} observed empirically in deep LLM layers, where the attention mechanism loses its discriminative ability among tokens and instead performs a near-uniform averaging across the sequence~\citep{rank3}. Gate-Norm makes this behavior predictable directly from the pre-trained query--key weights: as $m_\ell \to 0$, Equation~\ref{eq:softmax_uniform} drives $A_{ij}\to 1/S$ without any calibration data or forward passes.

\paragraph{3. From Uniformity to a Bounded Update:} 
Under near-uniform attention, the output of token $i$ decomposes as $\mathrm{AttnOut}_{\ell,i} = u_i + \Delta_i$, where $u_i$ is the sequence-mean value vector (set by $W_{v,\ell}$, $W_{o,\ell}$ and the activations) and $\Delta_i$ is the token-selective term, with $\|\Delta_i\| = O(m_\ell)$. To second order in $\|\mathrm{AttnOut}_{\ell,i}\|/\|X_{\ell,i}\|$, the cosine deviation is bounded by
\begin{equation}
1 - \cos(X_{\ell,i}, Y_{\ell,i}) \le \frac{\|u_i\|^2 + C\,m_\ell^2}{\|X_{\ell,i}\|^2}.
\label{eq:imp_bound}
\end{equation}
Gate-Norm controls $\Delta_i$ but not the floor $\|u_i\|$, so a small Gate-Norm alone does not guarantee a small output. We treat $\|u_i\|$ as empirical: Figure~\ref{fig:attention-collapse} shows that the total output $u_i+\Delta_i$ is small in deep layers, and Gate-Norm ranks layers by its token-selective part.

\paragraph{One-Shot Pruning Procedure:} 
GatePrune computes Gate-Norms across all layers, ranks them, and removes Self-attention layers with the smallest scores as shown in Algorithm~\ref{alg:GatePrune}.

\begin{algorithm}[h]
\caption{GatePrune: One-Shot Data-Free Pruning}
\label{alg:GatePrune}
\begin{algorithmic}[1]
  \Require Query/key weights $\{W_{q,\ell},W_{k,\ell}\}_{\ell=1}^L$, prune count $N$
  \For{$\ell=1$ \textbf{to} $L$}
    \State $m_\ell \gets \bigl\|W_{q,\ell}W_{k,\ell}^\top\bigr\|_F$  \Comment{Compute Gate-Norm Proxy (Eq.~\ref{eq:gate_norm})}
  \EndFor
  \State $\pi \gets \mathrm{argsort}(m_1,\dots,m_L)$  \Comment{Ascending order}
  \For{$i=1$ \textbf{to} $N$}
    \State Bypass Self-attention at layer $\pi[i]$  \Comment{Set $\mathrm{AttnOut}_{\pi[i]}=0$}
  \EndFor
  \State \Return Pruned model
\end{algorithmic}
\end{algorithm}

% !TEX root = ../Main.tex
% ----------------------------------------------------------------------------
\section{Experiments}
\label{sec:experiments}
This section evaluates GatePrune, a method for pruning self-Attention layers in LLMs. The results will highlight that GatePrune offers comparable accuracy to existing data-driven methods and is $1000\times$ faster than these methods without relying on any calibration data to select which self-Attention layers to prune.

We evaluate GatePrune on two LLaMA‐13B variants (v1 and v2)~\citep{LLama-v1, LLama-v2}, two Vicuna variants (7B and 13B)~\cite{vicuna2023} and LLaMA‐3.1-8B~\cite{Llama_3}, comparing against:  
(1)~\emph{ShortGPT}~\citep{shortgpt}, which drops entire Transformer blocks based on cosine similarity importance scores (Equation~\ref{eq:block_imp});  
(2)~\emph{Data‐driven Self-attention pruning}~\citep{not_all_attention}, which removes individual Self-attention layers via cosine similarity importance scores (Equation~\ref{eq:importance_scores});  
(3)~\emph{Random block removal}, which randomly selects and drops Transformer blocks; and  
(4)~\emph{Random Self-attention removal}, which randomly selects and removes Self-attention layers. \revised{All reported numbers are averaged over 3 runs.}

We measure:  
\textbf{Perplexity} on WikiText-2 as layers are removed (Section~\ref{subsec:perplexity}),  
\textbf{Zero-shot accuracy} on standard benchmarks (Section~\ref{subsec:zero_shot}), and \textbf{Pruning time}, i.e.\ time to compute importance scores and prune Self-attention layers (Section~\ref{subsec:efficiency}). 

\textbf{Hardware:} NVIDIA RTX A6000 GPU (48 GB, 10 752 CUDA cores, 336 Tensor cores, 309 TFLOPS peak) paired with an AMD EPYC 7713P 64-core CPU.  
\textbf{Software:} Ubuntu 20.04.6 LTS, Python 3.8.10, PyTorch 2.1.0.  
% ----------------------------------------------------------------------------
\subsection{Language Modeling Performance}
\label{subsec:perplexity}

Perplexity is evaluated on WikiText-2~\citep{wikitext} as the Self-attention layers are progressively removed. For both LLaMA-13B v1 and v2, we prune 1 to 25 layers in steps of 3 across three methods: Data‐driven Self-attention pruning, GatePrune, and random Self-attention removal. To ensure a fair comparison, we restrict this analysis to methods that only remove Self-attention layers, thereby excluding the block-level approach of ShortGPT (which we do include in our zero-shot accuracy evaluation in Section~\ref{subsec:zero_shot}).
Figure~\ref{fig:perplexity_plots} plots WikiText-2 perplexity versus number of dropped layers for v1 (left column) and v2 (right column), with full-range curves in the top row and zoomed-in comparisons in the bottom row. 

In \textsc{LLaMA}-13B v1, the baseline WikiText-2 perplexity is nearly 5.10. Removing 4--7 layers increases perplexity under data-driven pruning (5.25--5.46), and GatePrune closely follows (e.g., at 7 layers: 5.37 vs.\ 5.46). With 10--16 layers removed, perplexity grows gradually (data-driven from 6.3 to 8.6; GatePrune from 5.60 to 6.93), with GatePrune staying near or below the data-driven approach. Beyond 16 layers, perplexity rises sharply for both Self-attention-based methods. Meanwhile, random Self-attention removal is catastrophic (perplexity in the hundreds to thousands). % while block removal is intermediate---already worse than attention pruning at moderate budgets and degrading faster at larger budgets.

\textsc{LLaMA}-13B v2 shows the same pattern: the baseline is about 4.92 and increases modestly under moderate pruning (data-driven 5.14--5.28), with GatePrune closely following or marginally outperforming. Random Self-attention removal causes large increases in perplexity.

\textbf{Key Observations (Perplexity)}: On WikiText-2, GatePrune matches data-driven pruning on both \textsc{LLaMA}-13B variants across various pruning budgets, and outperforms a depth-only baseline that drops the last $N$ layers at every budget (Appendix~\ref{app:last_n}).

% 2-by-2 groupplot at 5.2cm per panel is far wider than one column: span both.
\begin{figure*}[t]
  \centering
  \begin{tikzpicture}
    \begin{groupplot}[
        group style={
          group size=2 by 2,
          horizontal sep=1cm,
          vertical sep=0.3cm        % reduced vertical separation
        },
        width=6.5cm,
        height=0.13\textwidth,
        scale only axis,
        xtick={1,4,7,10,13,16,19,22,25},
        tick label style={font=\footnotesize},
        label style={font=\footnotesize},
        grid style=dashed,
        ymajorgrids=true,
        legend style={font=\scriptsize, at={(rel axis cs:0,1)}, anchor=north west},
        legend cell align=left
      ]

      % ------------------------------------------------
      % Top-left: LLaMA-13B v1 full-range (no x tick labels)
      % ------------------------------------------------
      \nextgroupplot[
        title={\small LLaMA-13B v1},
        ylabel={Perplexity},
        xmin=1, xmax=25,
        xticklabels={},        % hide x tick labels
        xlabel={},             % no x-axis label
      ]
      % Data-driven importance
      \addplot[color=blue, mark=o, thick] coordinates {
        (1, 5.1022)
        (4, 5.2469)
        (7, 5.4593)
        (10, 6.3082)
        (13, 6.9982)
        (16, 8.6283)
        (19, 11.0122)
        (22, 16.3214)
        (25, 33.4299)
      };
      \addlegendentry{Data-driven}
      % GatePrune
      \addplot[color=orange, mark=square, thick] coordinates {
        (1, 5.1711)
        (4, 5.2875)
        (7, 5.3688)
        (10, 5.5961)
        (13, 6.3114)
        (16, 6.9337)
        (19, 10.2654)
        (22, 16.0330)
        (25, 33.5149)
      };
      \addlegendentry{GatePrune}
      % Random removal
      \addplot[color=green, mark=triangle, thick] coordinates {
        (1, 5.3981)
        (4, 5.9456)
        (7, 8.4375)
        (10, 9.8796)
        (13, 17.7457)
        (16, 23.3488)
        (19, 25.4635)
        (22, 569.5954)
        (25, 1322.0449)
      };
      \addlegendentry{Random}

      % ------------------------------------------------
      % Top-right: LLaMA-13B v2 full-range (no x tick labels)
      % ------------------------------------------------
      \nextgroupplot[
        title={\small LLaMA-13B v2},
        ylabel={},             % no repeated ylabel
        xmin=1, xmax=25,
        xticklabels={},        % hide x tick labels
        xlabel={}              % no x-axis label
      ]
      % Data-driven
      \addplot[color=blue, mark=o, thick] coordinates {
        (1, 4.9214)
        (4, 5.0618)
        (7, 5.1428)
        (10, 5.2810)
        (13, 5.7562)
        (16, 7.6451)
        (19, 11.9767)
        (22, 22.3865)
        (25, 60.3132)
      };
      \addlegendentry{Data-driven}
      % GatePrune
      \addplot[color=orange, mark=square, thick] coordinates {
        (1, 4.9179)
        (4, 4.9990)
        (7, 5.1283)
        (10, 5.3022)
        (13, 5.8033)
        (16, 8.2238)
        (19, 10.5059)
        (22, 16.3731)
        (25, 52.0042)
      };
      \addlegendentry{GatePrune}
      % Random
      \addplot[color=green, mark=triangle, thick] coordinates {
        (1, 5.2210)
        (4, 17.1223)
        (7, 25.7888)
        (10, 49.1984)
        (13, 79.9866)
        (16, 138.4600)
        (19, 147.7250)
        (22, 236.3390)
        (25, 370.7911)
      };
      \addlegendentry{Random}

      % ------------------------------------------------
      % Bottom-left: LLaMA-13B v1 zoomed (no title)
      % ------------------------------------------------
      \nextgroupplot[
        xlabel={Number of Self-attention layers dropped (N)},
        ylabel={Perplexity},
        xmin=1, xmax=25
        % automatic y-scaling
      ]
      % Data-driven
      \addplot[color=blue, mark=o, thick] coordinates {
        (1, 5.1022)
        (4, 5.2469)
        (7, 5.4593)
        (10, 6.3082)
        (13, 6.9982)
        (16, 8.6283)
        (19, 11.0122)
        (22, 16.3214)
        (25, 33.4299)
      };
      \addlegendentry{Data-driven}
      % GatePrune
      \addplot[color=orange, mark=square, thick] coordinates {
        (1, 5.1711)
        (4, 5.2875)
        (7, 5.3688)
        (10, 5.5961)
        (13, 6.3114)
        (16, 6.9337)
        (19, 10.2654)
        (22, 16.0330)
        (25, 33.5149)
      };
      \addlegendentry{GatePrune}

      % ------------------------------------------------
      % Bottom-right: LLaMA-13B v2 zoomed (no title)
      % ------------------------------------------------
      \nextgroupplot[
        xlabel={Number of Self-attention layers dropped (N)},
        ylabel={},
        xmin=1, xmax=25
        % automatic y-scaling
      ]
      % Data-driven
      \addplot[color=blue, mark=o, thick] coordinates {
        (1, 4.9214)
        (4, 5.0618)
        (7, 5.1428)
        (10, 5.2810)
        (13, 5.7562)
        (16, 7.6451)
        (19, 11.9767)
        (22, 22.3865)
        (25, 60.3132)
      };
      \addlegendentry{Data-driven}
      % GatePrune
      \addplot[color=orange, mark=square, thick] coordinates {
        (1, 4.9179)
        (4, 4.9990)
        (7, 5.1283)
        (10, 5.3022)
        (13, 5.8033)
        (16, 8.2238)
        (19, 10.5059)
        (22, 16.3731)
        (25, 52.0042)
      };
      \addlegendentry{GatePrune}

    \end{groupplot}
  \end{tikzpicture}
  \caption{
WikiText-2 Perplexity vs.\ number of dropped Self-attention layers on LLaMA-13B v1 (left) and v2 (right), up to 25 layers.
}
  \label{fig:perplexity_plots}
\end{figure*}

% LoRA fine-tuning figure (Figure~\ref{fig:lora_finetune}, discussed in the
% LoRA subsection below). Declared here, right after the perplexity figure, so
% LaTeX encounters it early enough to place it at the top of the next page
% alongside that figure. Declared later (after the zero-shot table*) it is
% deferred past the References into the appendix pages.
\begin{figure*}[t]
  \centering
  \begin{tikzpicture}
    \begin{groupplot}[
        group style={
          group size=2 by 1,
          horizontal sep=1.4cm,
        },
        width=6.6cm,
        height=0.15\textwidth,
        scale only axis,
        ymin=0,
        symbolic x coords={Unpruned, Pruned, {+LoRA}},
        xtick=data,
        enlarge x limits=0.4,
        tick label style={font=\footnotesize},
        label style={font=\footnotesize},
        grid style=dashed,
        ymajorgrids=true,
        nodes near coords={\pgfmathprintnumber[fixed, fixed zerofill, precision=1]{\pgfplotspointmeta}},
        every node near coord/.append style={font=\tiny, anchor=south, inner sep=1pt},
        legend style={font=\scriptsize, at={(rel axis cs:0,1)}, anchor=north west, draw=none, fill=none},
        legend cell align=left,
      ]

      % Left: Vicuna-13B (20 of 40 Self-attention layers pruned)
      \nextgroupplot[
        ybar=3pt, bar width=13pt,
        title={\small Vicuna-13B (20/40)},
        ylabel={Perplexity},
        ymax=32,
      ]
      \addplot[fill=orange!70, draw=orange!80!black] coordinates {
        (Unpruned, 5.95) (Pruned, 14.13) ({+LoRA}, 6.57)
      };
      \addlegendentry{GatePrune}
      \addplot[fill=blue!45, draw=blue!70!black] coordinates {
        (Unpruned, 5.95) (Pruned, 25.88) ({+LoRA}, 6.65)
      };
      \addlegendentry{Data-driven}

      % Right: LLaMA-2-13B (20 of 40 Self-attention layers pruned)
      \nextgroupplot[
        ybar=3pt, bar width=13pt,
        title={\small LLaMA-2-13B (20/40)},
        ylabel={},
        ymax=18,
      ]
      \addplot[fill=orange!70, draw=orange!80!black] coordinates {
        (Unpruned, 4.88) (Pruned, 14.11) ({+LoRA}, 6.35)
      };
      \addplot[fill=blue!45, draw=blue!70!black] coordinates {
        (Unpruned, 4.88) (Pruned, 14.73) ({+LoRA}, 6.27)
      };

    \end{groupplot}
  \end{tikzpicture}
  \caption{
WikiText-2 perplexity for Vicuna-13B (left) and LLaMA-2-13B (right) with 20 of 40 Self-attention layers pruned (\emph{Pruned}), then after brief LoRA fine-tuning (\emph{+LoRA}). Pruning sharply increases perplexity, and LoRA recovers it close to the unpruned baseline for both GatePrune (orange) and Data-driven (blue) pruning.
}
  \label{fig:lora_finetune}
\end{figure*}

\paragraph{Additional models.}
We further evaluate GatePrune on Vicuna-7B~\citep{vicuna2023}, Vicuna-13B~\citep{vicuna2023}, and LLaMA-3.1-8B~\citep{Llama_3}. We include LLaMA-3.1 because it utilizes Grouped Query Attention (GQA) rather than standard Multi-Head Attention, demonstrating the generality of GatePrune across different attention mechanisms~\citep{GQA}.
For each model, we progressively remove 1, 4, 7, 10, and 13 Self-attention layers. As summarized in Table~\ref{tab:additional_ppl}, random pruning adversely impacts perplexity once pruning becomes even moderately aggressive, whereas both GatePrune and the data-driven approach maintain low perplexity across a wide range of pruning budgets. GatePrune offers perplexity comparable to or lower than the data-driven baseline on all three models, and remains comparable at contexts up to 64K tokens (Appendix~\ref{app:long_context}).

% in the preamble you already have:
% \usepackage{xcolor}
% \usepackage{booktabs}

\begin{table}[t]
    \centering
    \caption{WikiText-2 perplexity on additional models as Self-attention layers are progressively pruned. \textbf{Bold} marks the better of the two importance-based methods (Data-driven vs.\ GatePrune) in each row; Random is an untrained baseline. Vicuna models use Multi-Head Attention (MHA); LLaMA-3.1-8B uses Grouped-Query Attention (GQA), so GatePrune generalizes across attention variants.}
    \label{tab:additional_ppl}
    \small
    \setlength{\tabcolsep}{4pt}
    \renewcommand{\arraystretch}{1.05}
    \resizebox{\columnwidth}{!}{%
    \begin{tabular}{@{}ccrrr@{}}
        \toprule
        \textbf{Model} & \textbf{\shortstack[c]{Layers\\pruned}} &
        \textbf{Random} & \textbf{\shortstack[c]{Data-\\driven}} &
        \textbf{GatePrune} \\
        \midrule
        \multirow{5}{*}{\shortstack[c]{\textbf{Vicuna-7B}\\[2pt]{\footnotesize (MHA, 32 heads)}}}
            & 1  & 6.82   & 7.23  & 7.23 \\
            & 4  & 7.84   & 7.62  & \textbf{7.42} \\
            & 7  & 13.84  & 8.78  & \textbf{8.57} \\
            & 10 & 355.87 & 12.55 & \textbf{10.02} \\
            & 13 & 473.40 & 20.46 & \textbf{16.64} \\
        \midrule
        \multirow{5}{*}{\shortstack[c]{\textbf{Vicuna-13B}\\[2pt]{\footnotesize (MHA, 40 heads)}}}
            & 1  & 6.00   & 5.99  & \textbf{5.97} \\
            & 4  & 10.63  & 6.11  & \textbf{6.05} \\
            & 7  & 67.09  & 6.24  & 6.24 \\
            & 10 & 587.90 & \textbf{6.39} & 6.47 \\
            & 13 & 594.83 & 7.54  & \textbf{7.05} \\
        \midrule
        \multirow{5}{*}{\shortstack[c]{\textbf{LLaMA-3.1-8B}\\[2pt]{\footnotesize (GQA, 32q\,/\,8kv heads)}}}
            & 1  & 6.51  & 6.35  & \textbf{6.30} \\
            & 4  & 8.11  & 6.83  & \textbf{6.67} \\
            & 7  & 14.59 & \textbf{7.39}  & 7.74 \\
            & 10 & 54.51 & \textbf{9.79}  & 11.55 \\
            & 13 & 98.55 & 16.48 & \textbf{15.65} \\
        \bottomrule
    \end{tabular}%
    }
\end{table}

% ----------------------------------------------------------------------------
\subsection{Zero‐Shot Accuracy Evaluation}
\label{subsec:zero_shot}

% 13 columns: must span both columns, otherwise \textwidth overflows the column.
\begin{table*}[t]
\centering
\scriptsize
\setlength{\tabcolsep}{3pt}
\caption{Zero-shot accuracies (\%) for LLaMA-13B variants.}
\label{tab:zero_shot_combined}
\vskip 0.1in
\resizebox{\linewidth}{!}{
\begin{tabular}{@{}l c c ccccccc ccc@{}}
\toprule
& & & \multicolumn{7}{c}{\textbf{Benchmark Tasks}} & \multicolumn{3}{c}{\textbf{Summary Metrics}} \\
\cmidrule(lr){4-10} \cmidrule(l){11-13}
\textbf{Method} & \textbf{\#Pruned} & \textbf{Type} & \textbf{BoolQ} & \textbf{RTE} & \textbf{HellaS.} & \textbf{WinoG.} & \textbf{ARC-E} & \textbf{ARC-C} & \textbf{OBQA} & \textbf{Avg.} & \textbf{Speed up} & \textbf{Data} \\ \midrule
\multicolumn{13}{c}{\textit{LLaMA-13B v1}} \\ \midrule
Baseline & 0 & -- & 79.85 & 68.95 & 79.32 & 73.40 & 77.36 & 48.38 & 47.40 & 67.81 & 1.00$\times$ & -- \\
\midrule
Random & 4 & Block & 68.62 & 54.15 & 71.30 & 64.80 & 72.94 & 41.13 & 40.20 & 59.02 & 1.10$\times$ & \xmark \\
Random & 8 & Block & 47.22 & 53.43 & 26.36 & 49.72 & 28.07 & 26.28 & 27.80 & 36.98 & 1.25$\times$ & \xmark \\
ShortGPT & 4 & Block & 63.49 & 68.23 & 76.60 & 72.77 & 74.03 & 45.99 & 46.20 & 63.90 & 1.10$\times$ & \cmark \\
ShortGPT & 8 & Block & 38.01 & 64.26 & 70.92 & 71.19 & 68.35 & 42.32 & 41.60 & 56.66 & 1.25$\times$ & \cmark \\
\midrule
Data-driven & 4 & Attn & 79.69 & 68.23 & 79.32 & 73.09 & 77.02 & 48.04 & 48.00 & 67.63 & 1.06$\times$ & \cmark \\
Data-driven & 8 & Attn & 79.69 & 61.73 & 78.82 & 73.16 & 76.85 & 47.87 & 48.00 & 66.59 & 1.12$\times$ & \cmark \\
Data-driven & 16 & Attn & 79.51 & 67.87 & 76.87 & 71.98 & 73.57 & 47.27 & 47.40 & 66.35 & 1.30$\times$ & \cmark \\
\midrule
\textbf{GatePrune} & 4 & Attn & 79.69 & 66.79 & 79.27 & 72.69 & 76.98 & 48.81 & 47.80 & 67.43 & 1.06$\times$ & \xmark \\
\textbf{GatePrune} & 8 & Attn & 79.79 & 63.90 & 79.06 & 72.61 & 76.35 & 47.10 & 47.40 & 66.60 & 1.12$\times$ & \xmark \\
\textbf{GatePrune} & 16 & Attn & 79.20 & 69.31 & 77.24 & 73.09 & 73.40 & 47.53 & 46.60 & 66.62 & 1.30$\times$ & \xmark \\
\midrule \midrule
\multicolumn{13}{c}{\textit{LLaMA-13B v2}} \\ \midrule
Baseline & 0 & -- & 82.11 & 67.51 & 79.62 & 72.61 & 78.96 & 48.81 & 45.20 & 67.83 & 1.00$\times$ & -- \\
\midrule
Random & 4 & Block & 62.42 & 53.07 & 52.80 & 53.43 & 50.25 & 29.01 & 29.60 & 47.23 & 1.10$\times$ & \xmark \\
Random & 8 & Block & 56.94 & 52.35 & 29.37 & 48.62 & 30.09 & 25.00 & 24.40 & 38.11 & 1.25$\times$ & \xmark \\
ShortGPT & 4 & Block & 78.72 & 65.70 & 77.12 & 72.14 & 75.93 & 47.44 & 45.60 & 66.09 & 1.10$\times$ & \cmark \\
ShortGPT & 8 & Block & 52.57 & 63.90 & 72.60 & 69.69 & 67.21 & 43.17 & 41.20 & 58.62 & 1.25$\times$ & \cmark \\
\midrule
Data-driven & 4 & Attn & 81.99 & 68.59 & 79.48 & 72.61 & 79.08 & 48.72 & 46.00 & 68.07 & 1.06$\times$ & \cmark \\
Data-driven & 8 & Attn & 81.99 & 66.43 & 79.39 & 72.30 & 78.54 & 48.21 & 46.40 & 67.61 & 1.12$\times$ & \cmark \\
Data-driven & 16 & Attn & 80.73 & 61.01 & 78.16 & 72.22 & 76.85 & 48.63 & 47.00 & 66.37 & 1.30$\times$ & \cmark \\
\midrule
\textbf{GatePrune} & 4 & Attn & 82.51 & 66.43 & 79.61 & 72.53 & 78.75 & 49.32 & 46.20 & 67.91 & 1.06$\times$ & \xmark \\
\textbf{GatePrune} & 8 & Attn & 81.99 & 62.09 & 79.69 & 72.69 & 78.32 & 48.46 & 46.40 & 67.09 & 1.12$\times$ & \xmark \\
\textbf{GatePrune} & 16 & Attn & 80.80 & 60.65 & 78.63 & 72.61 & 77.44 & 47.95 & 46.60 & 66.38 & 1.30$\times$ & \xmark \\
\bottomrule
\end{tabular}
}
\end{table*}

Zero-shot performance on seven NLP benchmarks (BoolQ~\citep{boolq}, RTE~\citep{RTE}, HellaSwag~\citep{Hellaswag}, WinoGrande~\citep{Winogrande}, ARC-Easy/Challenge~\citep{arc}, OpenBookQA~\citep{OpenbookQA}) is summarized in Table~\ref{tab:zero_shot_combined} for \textsc{LLaMA}-13B v1 and v2. 

The average accuracy of the unpruned baselines is 67.8\% for both v1 and v2. Pruning 4 Self-attention layers with GatePrune reduces average accuracy by at most 0.4 percentage points (on v2 it is within 0.1 points of the baseline), closely matching data-driven pruning. At 8 layers, both methods lose at most 1.2 points. Even with 16 of 40 layers removed, GatePrune stays within 1.5 points of the baseline (1.2 on v1 and 1.5 on v2) and matches data-driven pruning (66.62 vs.\ 66.35 on v1; 66.38 vs.\ 66.37 on v2).

For the inference speed--accuracy trade-off, removing 4 Self-attention layers with GatePrune yields a $1.06\times$ speed-up for at most a 0.4-point loss in average accuracy; 8 layers gives $1.12\times$ for at most 1.2 points; and 16 layers delivers $1.30\times$ for at most 1.5 points. Block removal (ShortGPT) shows substantially larger accuracy declines at comparable speed-ups: 1.7--3.9 points at $1.10\times$ and 9.2--11.2 points at $1.25\times$.

\textbf{Key Observation (Zero‐Shot Accuracy)}: GatePrune matches the zero-shot accuracy of data-driven pruning while delivering up to a $1.30\times$ inference speed-up.
\revised{\subsection{Post-pruning fine-tuning with LoRA}
We now explore the effect of brief fine-tuning after pruning. We prune half the Self-attention layers and perform fine-tuning using LoRA for 2000 steps on WikiText-2. We run this protocol on Vicuna-13B and LLaMA-2-13B (v2).}

Figure~\ref{fig:lora_finetune} reports WikiText-2 perplexities before pruning, after pruning 20 Self-attention layers, and after the subsequent LoRA fine-tuning. Pruning 20 layers without fine-tuning increases perplexity, whereas including LoRA recovers performance on both models. Notably, GatePrune matches or improves over data-driven pruning in terms of perplexity (specifically on Vicuna-13B). Overall, this indicates that GatePrune, when combined with brief fine-tuning, recovers perplexity close to that of the unpruned baseline while removing half of the Self-attention layers in LLaMA-2-13B and Vicuna-13B.

\subsection{End‐to‐End Pruning Efficiency}
\label{subsec:efficiency}

In a 40-layer, 13 B-parameter LLaMA model on an NVIDIA RTX A6000, GatePrune prunes Self-attention layers in nearly \textbf{300 ms}, delivering over \(\mathbf{1000\times}\) speed-up compared to data-driven pruning.

As GatePrune operates solely on pre‐trained weight matrices, it requires no forward passes. Data-driven pruning, by contrast, must load the full model and thousands of calibration tokens into GPU memory and run forward passes, sharding across GPUs when the model exceeds a single device. GatePrune can instead stream weight matrices into system RAM, pruning a 13 B‐parameter LLaMA checkpoint in under \textbf{30 s} on a CPU---even when the model does not fit in GPU memory. 

% !TEX root = ../Main.tex
\section{Discussion and Conclusion}

We presented the \emph{Attention Suppression Hypothesis}: as LLMs are trained, deeper Self-attention layers learn to suppress their own outputs, routing information through the residual connection into the following MLP. This is evidenced by the attention-to-input norm ratio \(r_\ell\), which falls sharply in later layers, indicating that beyond a certain depth Self-attention contributes little to the residual stream.

To this end, we introduce GatePrune, which identifies and removes redundant Self-attention layers using only pre-trained weights, without any forward passes or calibration data. It matches the accuracy and perplexity of data-driven pruning while accelerating pruning by $1000\times$, and its weight-only, CPU-friendly design makes it practical even for models larger than available GPU memory.

%In this work, we focus on dense decoder-only Transformers; extending such weight-only criteria to mixture-of-experts routing and expert modules is an interesting but distinct direction for future work.

% The Attention Suppression phenomenon and GatePrune together suggest that many pre-trained Transformer layers harbor inherent redundancy exploitable without additional data or fine-tuning. This opens avenues for architectural refinement—e.g., reducing or restructuring late-stage attention, dynamic depth adaptation during inference, or hybrid designs that allocate capacity more judiciously across layers. Future work may extend these ideas to other modalities (vision, multimodal transformers).

% ================================
% Limitations
% ================================
\section*{Limitations}
GatePrune opens two avenues for future work. First, it prunes only
Self-Attention layers, leaving the potential for considering MLP (feed-forward) sublayers;
a comparable weight-only metric for MLP pruning is a promising direction. Second, we
target dense Transformers, and extending Gate-Norm to Mixture-of-Experts
models, where sparse expert routing may not exhibit the same suppression
pattern, is an important next step.

% -----------------------
% References
% -----------------------
% NOTE: the style file already issues \bibliographystyle{acl_natbib}. Repeating it here
% writes a second \bibstyle into Main.aux, which makes BibTeX abort with
% "Illegal, another \bibstyle command" and leaves every citation undefined.
\bibliography{acl2027}   % your .bib file (rename if you like)

% -----------------------
% Appendix
% -----------------------
\appendix
% !TEX root = ../Main.tex
\section{Detailed Derivation of the Gate-Norm Bound}
\label{app:gate_norm_proofs}

In Section~\ref{sec:gate_norm}, we stated that the Gate-Norm \(m_\ell\) controls the token-selective part of the Self-attention output. Here we provide the step-by-step derivation of the bound
\begin{equation}
1-\cos(X_{\ell,i},Y_{\ell,i}) \;\le\; \frac{\|u_i\|^2 + C\,m_\ell^2}{\|X_{\ell,i}\|^2},
\end{equation}
where \(u_i\) is the sequence-mean value vector, which the Gate-Norm does not control.

\subsection{Setup and Notation}
The output of a single Self-attention layer and the residual stream is:
\begin{equation}
\begin{split}
Y_\ell &= X_\ell + \mathrm{AttnOut}_\ell, \\
\mathrm{Imp}_\ell^{\mathrm{attn}} &=1-\cos\bigl(X_\ell,Y_\ell\bigr).
\end{split}
\end{equation}
After normalization at layer \(\ell\), let the per-token embeddings be \(z_1,\dots,z_S\in\mathbb{R}^D\). For a single head, we define:
\begin{equation}
\begin{split}
W_{q,\ell},\,W_{k,\ell} &\in\mathbb{R}^{D\times d_h}, \\
M_\ell &= W_{q,\ell}W_{k,\ell}^\top, \\
m_\ell &= \|M_\ell\|_F.
\end{split}
\end{equation}

\subsection{Step 1: Logit Bound}
The raw Self-attention logits are:
\begin{equation}
L_{ij}
= \frac{1}{\sqrt{d_h}}\;z_i^\top M_\ell\,z_j.
\end{equation}
By Cauchy--Schwarz:
\begin{equation}
|L_{ij}|\le \frac{\|z_i\|\;\|M_\ell\|_F\;\|z_j\|}{\sqrt{d_h}}
\le C_0\,m_\ell,
\end{equation}
where \(C_0\) depends only on the token norms \(\|z_i\|\). Both LayerNorm and the RMSNorm used by LLaMA bound these norms (for RMSNorm with gain \(\gamma_\ell\), \(\|z_i\|\le\|\gamma_\ell\|_\infty\sqrt{D}\)). Hence:
\begin{equation}
\|L\|_\infty = O(m_\ell).
\end{equation}

\subsection{Step 2: Nearly Uniform Softmax}
When \(\|L\|_\infty = O(m_\ell)\ll1\), every weight in a row of the softmax lies within a factor \(e^{\pm 2C_0m_\ell}\) of \(1/S\), where \(S\) is the number of positions the row attends to (under causal masking, \(S=i\) for token \(i\)). Thus:
\begin{equation}
A_{ij}=\frac1S+\delta_{ij},\quad \sum_j\delta_{ij}=0,\quad \sum_j|\delta_{ij}|\le C_1\,m_\ell.
\end{equation}

\subsection{Step 3: Decompose the Self-attention Output}
Let \(V_jW_{o,\ell}\in\mathbb{R}^D\) be the values followed by the output projection. Then:
\begin{equation}
\begin{split}
\mathrm{AttnOut}_{\ell,i}
&=\sum_{j=1}^S A_{ij}\,(V_jW_{o,\ell}) \\
&=\underbrace{\frac1S\sum_j V_jW_{o,\ell}}_{u_i}
\;+\;\underbrace{\sum_j \delta_{ij}\,(V_jW_{o,\ell})}_{\Delta_i}.
\end{split}
\end{equation}
By Step~2,
\begin{equation}
\|\Delta_i\| \le\sum_j|\delta_{ij}|\cdot\|V_jW_{o,\ell}\|=O(m_\ell).
\end{equation}
The Gate-Norm controls only the token-selective term \(\Delta_i\). The sequence mean \(u_i\) is set by \(W_{v,\ell}\), \(W_{o,\ell}\) and the activations, and the Gate-Norm places no constraint on it. For example, \(W_{q,\ell}=0\) gives \(m_\ell=0\) and exactly uniform attention, but \(\|u_i\|\) can be large. We therefore keep \(\|u_i\|\) as an explicit term.

\subsection{Step 4: Bounding Cosine Deviation via second-Order Expansion}
The output of the Self-attention layer and the residual stream is \(Y_{\ell,i} = X_{\ell,i} + v_i\), where \(v_i = u_i + \Delta_i\). A second-order expansion of the cosine in \(\|v\|/\|x\|\) gives:
\begin{align}
&1 - \cos(x,\,x+v) \nonumber \\
&\quad =\; 1 - \frac{\|x\|^2 + x \cdot v}
              {\|x\| \sqrt{\|x\|^2 + 2x \cdot v + \|v\|^2}} \nonumber \\
&\quad \approx\; \frac{\|v\|^2}{2\|x\|^2} - \frac{(x \cdot v)^2}{2\|x\|^4}.
\end{align}
The subtracted term is the component of \(v\) parallel to \(x\). Thus, using only cosine similarity cannot detect a parallel update (Section~\ref{sec:active_suppression}):
\begin{equation}
1 - \cos(x,\,x+v) \;\le\; \frac{\|v\|^2}{2\|x\|^2}.
\end{equation}

\subsection{Step 5: The Final Bound}
Since \(\|u_i + \Delta_i\| \le \|u_i\| + \|\Delta_i\|\) and \((\|u_i\| + \|\Delta_i\|)^2 \le 2(\|u_i\|^2 + \|\Delta_i\|^2)\), we have \(\tfrac12\|u_i + \Delta_i\|^2 \le \|u_i\|^2 + \|\Delta_i\|^2\). Substituting this and Step~3 into Step~4 with \(x = X_{\ell,i}\) and \(v = u_i + \Delta_i\) gives:
\begin{equation}
1 - \cos\bigl(X_{\ell,i},\,Y_{\ell,i}\bigr)
\;\le\;
\frac{\|u_i\|^2 + C\,m_\ell^2}{\|X_{\ell,i}\|^2}.
\end{equation}

\subsection{Conclusion}
The first term, \(\|u_i\|^2\), is a floor that does not depend on \(m_\ell\), so a small Gate-Norm alone does not guarantee a small Self-attention output. We consider \(\|u_i\|\) as empirical. Appendix~\ref{app:correlation} shows that the Gate-Norm ranking closely tracks measured layer importance.

% !TEX root = ../Main.tex
\section{Additional Experiments}
\label{app:add_exp}

\subsection{Dropping the Last $N$ Layers}
\label{app:last_n}
Figure~\ref{fig:attention-collapse} shows that the attention-to-input norm ratio decays with depth. To test whether GatePrune's gains come only from depth, we compare it against a simple data-free baseline, \emph{Last-$N$}, which removes the $N$ deepest Self-attention layers. Table~\ref{tab:last_n} reports WikiText-2 perplexity on LLaMA-13B v2 when the $N$ deepest Self-attention layers are removed.

\begin{table}[h]
\centering
\caption{WikiText-2 perplexity on LLaMA-13B v2 when $N$ Self-attention layers are removed. Last-$N$ drops the $N$ deepest Self-attention layers.}
\label{tab:last_n}
\small
\setlength{\tabcolsep}{3pt}
\resizebox{\columnwidth}{!}{%
\begin{tabular}{@{}lrrrrrrrr@{}}
\toprule
$N$ & 4 & 7 & 10 & 13 & 16 & 19 & 22 & 25 \\
\midrule
Last-$N$    & 6.34 & 7.18 & 7.68 & 8.17 & 10.26 & 14.69 & 22.39 & 60.31 \\
Data-driven & 5.06 & 5.14 & 5.28 & 5.76 & 7.65  & 11.98 & 22.39 & 60.31 \\
GatePrune   & 5.00 & 5.13 & 5.30 & 5.80 & 8.22  & 10.51 & 16.37 & 52.00 \\
\bottomrule
\end{tabular}%
}
\end{table}

GatePrune outperforms Last-$N$ at every $N$, so its gains are not only due to depth. At $N\ge22$, data-driven pruning also selects the last $N$ layers, so the two coincide.

\paragraph{Selected layers.} GatePrune does not simply select the deepest layers. On LLaMA-13B v2 with $N=16$, it removes Self-attention layers $\{23\text{--}36, 38, 39\}$ (indexed 1--40), retaining layers 37 and 40.

\subsection{Long-Context Evaluation}
\label{app:long_context}
GatePrune removes entire Self-attention layers, so a layer that is redundant on short inputs could become important when long-range token interactions are required. We therefore measure perplexity on PG-19~\citep{pg19}, a corpus of full-length books, at context lengths from 4K to 64K tokens on LLaMA-3.1-8B (Table~\ref{tab:long_context}).

\begin{table}[h]
\centering
\caption{PG-19 perplexity on LLaMA-3.1-8B by context length, with $N$ Self-attention layers removed.}
\label{tab:long_context}
\small
\setlength{\tabcolsep}{4pt}
\resizebox{\columnwidth}{!}{%
\begin{tabular}{@{}lcrrrrr@{}}
\toprule
Method & $N$ & 4K & 8K & 16K & 32K & 64K \\
\midrule
Dense       & 0 & 10.07 & 9.65  & 9.35  & 9.13  & 8.97  \\
\midrule
Data-driven & 1 & 10.17 & 9.75  & 9.45  & 9.22  & 9.06  \\
GatePrune   & 1 & 10.17 & 9.75  & 9.45  & 9.22  & 9.06  \\
\midrule
Data-driven & 4 & 10.65 & 10.22 & 9.91  & 9.69  & 9.53  \\
GatePrune   & 4 & 10.65 & 10.22 & 9.91  & 9.69  & 9.53  \\
\midrule
Data-driven & 7 & 11.59 & 11.15 & 10.83 & 10.60 & 10.45 \\
GatePrune   & 7 & 12.08 & 11.68 & 11.41 & 11.24 & 11.13 \\
\bottomrule
\end{tabular}%
}
\end{table}

At $N=1$ and $N=4$, GatePrune and data-driven pruning select the same layers, so their perplexities are identical. At $N=4$, the increase over the dense model is nearly constant as the context grows, so the removed layers remain redundant at long context lengths. At $N=7$, GatePrune trails data-driven pruning, consistent with the short-context results in Table~\ref{tab:additional_ppl} (7.74 vs.\ 7.39), and the gap widens with increasing context length, from 0.49 at 4K to 0.68 at 64K. GatePrune therefore achieves comparable perplexity to data-driven pruning at long context lengths without using any calibration data.

\subsection{Effect of Removal on the Block Output}
\label{app:block_output}
A small attention-to-input norm ratio $r_\ell$ only does not guarantee that a Self-attention layer can be removed safely. The MLP that follows receives the altered input and could amplify the change. We therefore measure the change in the block output directly. On LLaMA-13B v1, we remove the Self-attention layer of one block at a time, pass the result through the MLP, and compute the cosine similarity between the original and pruned block outputs on C4 (Table~\ref{tab:block_output}).

\begin{table}[h]
\centering
\caption{Attention-to-input norm ratio $r_\ell$ and cosine similarity between the original and pruned block outputs on LLaMA-13B v1 (C4). $^{*}$ marks layers removed by GatePrune at $N=16$.}
\label{tab:block_output}
\small
\setlength{\tabcolsep}{4pt}
\resizebox{\columnwidth}{!}{%
\begin{tabular}{@{}lccccccc@{}}
\toprule
Layer & 10 & 15 & 20$^{*}$ & 25$^{*}$ & 30$^{*}$ & 35$^{*}$ & 40 \\
\midrule
$r_\ell$    & 0.33  & 0.27  & 0.19  & 0.12  & 0.08  & 0.08  & 0.11  \\
Cosine sim. & 0.947 & 0.966 & 0.983 & 0.993 & 0.997 & 0.996 & 0.980 \\
\bottomrule
\end{tabular}%
}
\end{table}

Layers with a small $r_\ell$ leave the block output nearly unchanged. Every layer that GatePrune removes has a cosine similarity of at least 0.983, so the MLP does not amplify the change. Layer 40 is an exception: despite a small $r_\ell$, its similarity is lower (0.980), and GatePrune retains it.

\subsection{Correlation Between Gate-Norm and Layer Importance}
\label{app:correlation}
GatePrune ranks layers by only using Gate-Norm. To test whether this ranking reflects actual layer importance, we compute, for each of the 40 Self-attention layers of LLaMA-13B v1, the Gate-Norm $m_\ell$ and three measures of importance: (i) the attention-to-input norm ratio $r_\ell$; (ii) the cosine-based importance $\mathrm{Imp}_\ell^{\mathrm{attn}}$ (Equation~\ref{eq:importance_scores}); and (iii) the increase in WikiText-2 perplexity when only that layer's Self-attention is removed. The importance measures (i) and (ii) are computed on the calibration set used for data-driven pruning (128 sequences of 2048 tokens from C4). Table~\ref{tab:correlation} reports the Spearman rank correlation between $m_\ell$ and each measure.

\begin{table}[h]
\centering
\caption{Spearman rank correlation between Gate-Norm $m_\ell$ and layer importance across the 40 Self-attention layers of LLaMA-13B v1.}
\label{tab:correlation}
\small
\begin{tabular}{@{}lc@{}}
\toprule
Importance measure & Spearman $\rho$ \\
\midrule
(i) Norm ratio $r_\ell$                          & 0.91 \\
(ii) Cosine importance $\mathrm{Imp}_\ell^{\mathrm{attn}}$ & 0.91 \\
(iii) Single-layer ablation ($\Delta$PPL)        & 0.85 \\
\bottomrule
\end{tabular}
\end{table}

Ranking using Gate-Norm is strongly correlated to all three measures, including the increase in perplexity when removing a layer. A ranking computed only from the query and key weights therefore closely tracks the importance measured with forward passes.

\end{document}